\documentclass[10pt,twocolumn,letterpaper]{article}

\PassOptionsToPackage{table}{xcolor}
\usepackage[pagenumbers]{cvpr} 

\usepackage{adjustbox}
\usepackage{cuted} 
\usepackage{booktabs,graphicx, tabularx, multirow, colortbl} 
\usepackage{xcolor}          
\usepackage{amsmath,amssymb}        
\usepackage{bm}                     
\usepackage{graphicx}               
\usepackage{algorithm}              
\usepackage{algpseudocode}          
\usepackage{threeparttable}
\usepackage{makecell}
\usepackage{pifont}
\usepackage{xfrac}
\usepackage{amsmath}
\usepackage{placeins}

\newcommand{\cmark}{\ding{51}} 
\newcommand{\xmark}{\ding{55}} 

\definecolor{cvprblue}{rgb}{0.21,0.49,0.74}
\usepackage[pagebackref,breaklinks,colorlinks,allcolors=cvprblue]{hyperref}

\def\confName{CVPR}
\def\confYear{2027}

\title{Xtream: An Explicit Metric State for Large-Scale Streaming Reconstruction}

\author{
Ernesto Lozano \quad
Alberto Jaenal \quad
Javier Civera \\[2mm]
I3A, Universidad de Zaragoza, Spain\\
{\tt\small \{e.lozano, ajaenal, jcivera\}@unizar.es}
}

\begin{document}


\twocolumn[{%
  \renewcommand\twocolumn[1][]{#1}%
  \maketitle
  \begin{center}
    \centering
    \captionsetup{type=figure}
    \includegraphics[width=0.96\textwidth]{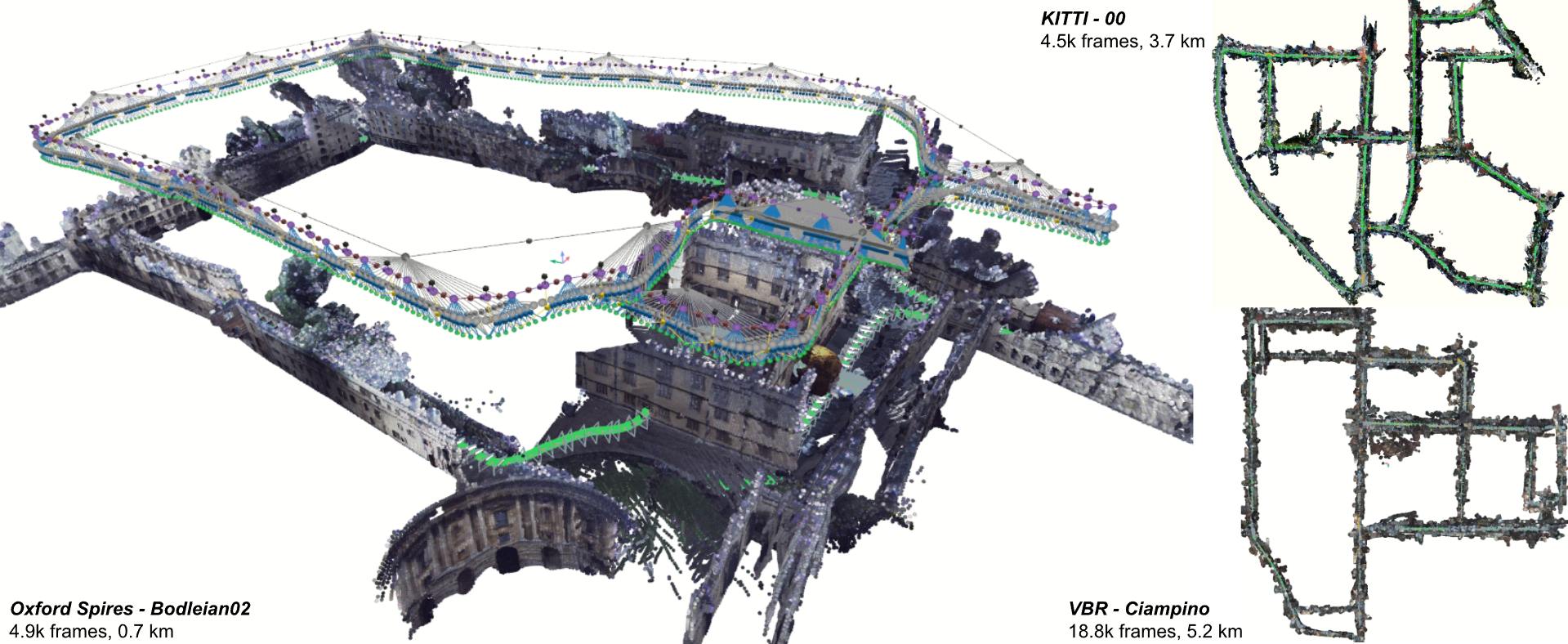}
    \captionof{figure}{\textbf{Camera trajectories and scene reconstructions using Xtream.} Left: Reconstruction from the Oxford Spires dataset. We also show the underlying factor graph, where our main contributions are. Right: Top view of reconstructions from the KITTI and VBR datasets.}
    \label{fig:teaser}
  \end{center}%
  \vspace{0.6em} 
}]

\begin{abstract}
3D foundation models (3DFMs) enable dense reconstruction in a feed-forward pass, yet their effective span is limited by cross-view attention. Local consistency has been expanded through geometric alignment of video chunks and recursive updates of an implicit memory, and large-scale scenes with pose graphs. However, both strategies are prone to drift, largely due to the lack of metric-scale observability from monocular geometry. We present Xtream, a hybrid streaming pipeline that maintains an explicit metric state through a factor graph over keyframes, while using a 3DFM as a geometric front end.
Xtream incorporates metric information in two ways: through metric scale predictions from 3DFMs; or through inertial measurements if available.
Across the large-scale benchmarks of VBR, Oxford Spires and KITTI, Xtream achieves state-of-the-art performance in 3DFM-based online reconstruction, reducing average absolute trajectory errors compared with published baselines and concurrent work.
\small{\url{https://ernestolozano.github.io/xtream}}
\end{abstract}    
\vspace{-2em}
\section{Introduction}
\label{sec:intro}

Traditionally, 3D reconstruction approaches have relied on feature-based front ends and local optimization back ends that exploit multi-view constraints~\cite{qin2018vinsmono,schoenberger2016sfm,campos2021orbslam3,boche2025okvis2x,pan2024global}. However, these approaches, struggle in scenarios with limited visual texture or poorly constrained geometric configurations.
3DFMs~\cite{wang2024dust3r,leroy2024mast3r,wang2025vggt,lin2025da3} provide an alternative. Trained on large-scale databases, they predict dense geometry directly from images, even under challenging texture and viewing conditions.
However, the computational cost of cross-view attention limits the number of views that 3DFMs can process jointly. To handle long image sequences in an online setting, existing approaches follow two main strategies: either they process overlapping fixed-size windows (or chunks) and enforce geometric consistency across them~\cite{murai2025mast3rslam, maggio2026vggtslam,deng2026vggtlong,wang2026amb3r,wang2026amb3rslam}, or they maintain local consistency through an implicit memory~\cite{wang2025cut3r,zhuo2025streamvggt,chen2026ttt3r,zhang2026loger,li2026closer}. To further enforce global consistency, several approaches additionally employ an explicit graph that models geometric relationships between selected keyframes~\cite{murai2025mast3rslam, maggio2026vggtslam,li2026closer,deng2026vggtlong,wang2026amb3rslam}.

While recent work has demonstrated impressive results, we identify monocular scale drift as a major source of error in 3DFM-based SLAM and streaming methods. Even when loop closures are detected, local optimization may be insufficient to recover global geometric consistency in the presence of substantial scale drift. We propose to leverage metric-scale predictions from scale-aware 3DFMs~\cite{lin2025da3,wang2025pi3,keetha2026mapanything}, or metric constraints from inertial measurement units (IMUs), to mitigate scale drift and improve the geometric consistency of 3DFM-based streaming reconstruction.

We present {Xtream}, a hybrid streaming pipeline that builds on 3DFMs and factor graphs for large-scale reconstruction. Following previous work, we represent the scene as a set of keyframes, and treat the 3DFM as a front end that provides relative constraints. In addition to adding constraints between consecutive frames, we also detect co-visibility with a visual place recognition model, and add constraints between separate keyframes and loop closures. 

\begin{figure*}[!ht]
  \centering
  \includegraphics[width=\textwidth]{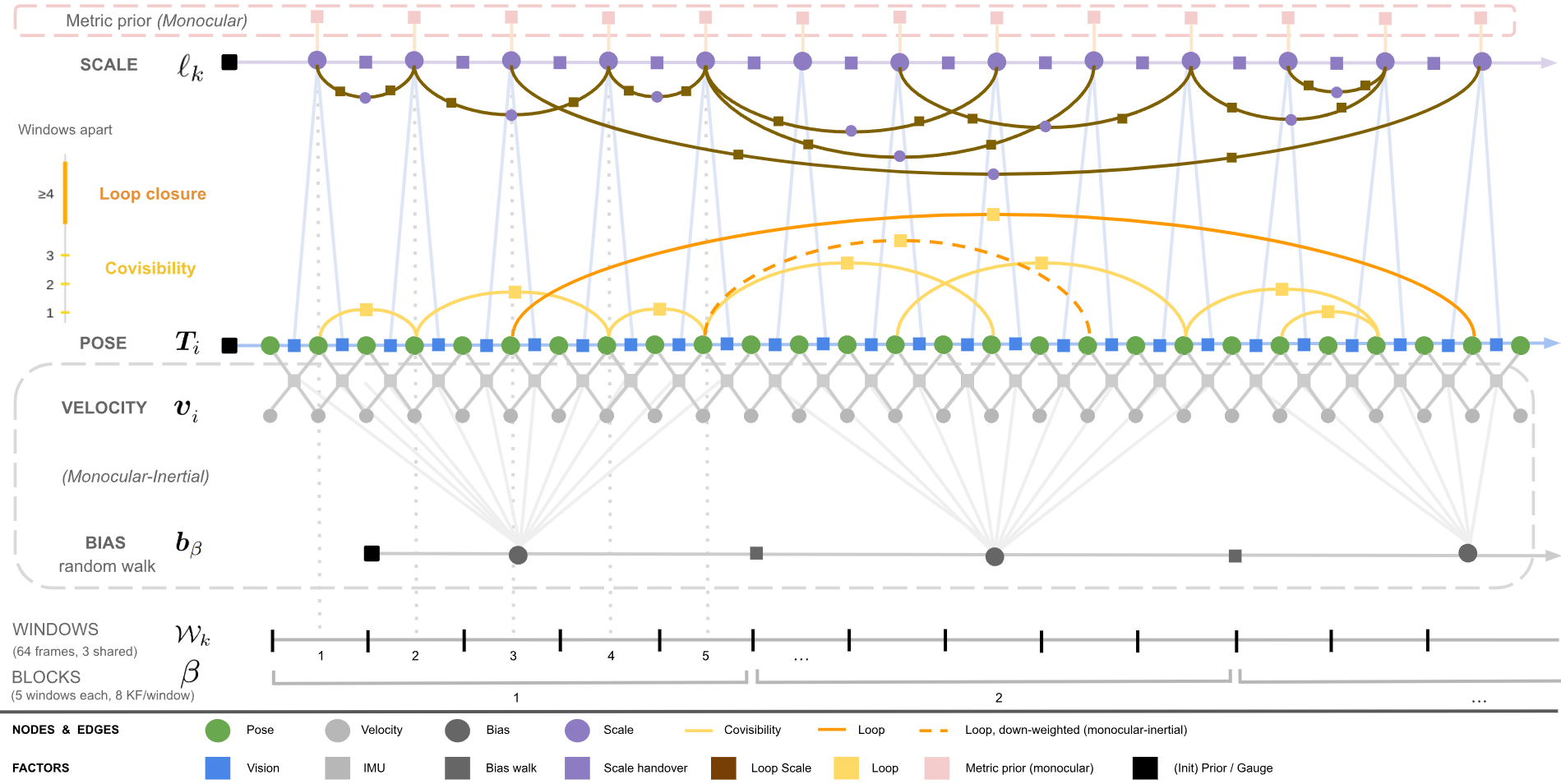}
  \caption{\textbf{Overview of Xtream's pose graph.} Per-window metric scale, plotted as purple nodes in the upper part, are constrained by 3DFM's metric priors (pink factors) in the monocular case, and by inertial constraints in the mono-inertial case (gray factors). Scale is also constrained between consecutive (purple factors) and distant (brown factors) windows, and by visual residuals (blue factors).}
  \label{fig:overview}
\end{figure*}

Our specific contributions are:
\begin{enumerate}
  \item An \textbf{explicit model of per-window metric scale}, for monocular and mono-inertial settings. See them illustrated as purple and brown nodes and edges in \cref{fig:overview}.
  \item \textbf{Explicit formulations of absolute metric scale residuals}, from inertial readings if an IMU is present, or from metric 3DFMs. See them illustrated, respectively, as grey and pink nodes in the lower and upper part of \cref{fig:overview}.
  \item \textbf{The integration} of these metric scale models into our full 3DFM-based streaming pipeline, Xtream.
\end{enumerate}

Our ablation in five different benchmarks confirm that metric scale modeling plays a major role in reducing trajectory errors. In our experimental comparisons, Xtream outperforms published and concurrent baselines in kilometer-scale benchmarks (VBR, Oxford Spires and KITTI), while being competitive in meter-scale benchmarks (EuRoC and ScaRF-SLAM), in which scale drift is less pronounced.
\section{Related work}
\label{sec:related_work}

Following the emergence of 3DFMs~\cite{wang2024dust3r,leroy2024mast3r,wang2025vggt,wang2025pi3,lin2025da3,wang2026vggt-O,keetha2026mapanything}, several methods have extended their limited multi-view context to long sequences.
Some of them propose architectural changes for online sequence processing~\cite{cabon2025must3r,chen2026ttt3r,chen2025long3r,liu2025slam3r,taher2026kv,zhang2026loger,lan2025stream3r,li2025wint3r,wang2025cut3r,khafizov2025gcut3r,mahdi2025evict3r,wu2025point3r,ma2025puzzles,yuan2026slamformer,zhuo2026streaming,chen2026lingbot}, using implicit memory modules such as recurrent memory, fast weights from test-time or key-value cache.
Other approaches divide long videos into overlapping windows or chunks~\cite{deng2026vggtlong,wang2025pi3,lin2025da3}, aligning them afterwards.

Inspired by classical SLAM approaches~\cite{campos2021orbslam3,boche2025okvis2x,qin2018vinsmono}, several works propose a hybrid approach, adding a graph of keyframe poses to the aligned 3DFM windows. MASt3R-SLAM~\cite{murai2025mast3rslam} optimizes in the back end a graph of Sim(3) camera poses to account for scale differences, similarly to AMB3R-SLAM~\cite{wang2026amb3rslam}, concurrent to our work, that shows the importance of maintaining such graph. The VGGT-SLAM series~\cite{maggio2026vggtslam,maggio2025vggt-slam2} optimize in the SL(4) manifold, accounting for the degrees of freedom of uncalibrated cameras. CLoSeR~\cite{li2026closer}, however, proposes to optimize SE(3) poses, arguing that their base 3DFM LoGeR~\cite{zhang2026loger} achieves better scale consistency than MASt3R~\cite{leroy2024mast3r} and VGGT~\cite{wang2025vggt}, used by previous works. In Xtream, mostly based in but not restricted to Depth Anything 3~\cite{lin2025da3}, we show that scale drift keeps being the largest source of monocular drift, and that our explicit models of metric states and metric measurements yields state-of-the-art results. 

Recent work has integrated inertial cues into 3DFM-based pipelines, motivated by the robustness and accuracy of classical visual-inertial pipelines~\cite{qin2018vinsmono,campos2021orbslam3,boche2025okvis2x}. \citet{zhang2026efficient} propose a closed-form initialization for visual-inertial SLAM pipelines using 3DFM outputs. In contrast, VI3~\cite{lozano2026vi3} adds metric scale to 3D reconstructions by combining IMU preintegration with 3DFM predictions.
MASt3R-Fusion~\cite{zhou2025mast3rfusion} proposes an online factor graph merging MASt3R two-view correspondences with IMU measurements to recover metric scale. 
FoundationVIO~\cite{rhee2026foundationvio}, developed concurrently with our work, is a 3DFM-based visual-inertial pipeline targeting long-term consistency and metric-scale recovery. We adopt its relative-scale parameterization and extend it to incorporate metric-scale constraints from 3DFMs. Furthermore, unlike FoundationVIO, which relies on partial trajectories for alignment, we use point clouds, resulting in more accurate alignment. These contributions enable Xtream to outperform all the aforementioned visual-inertial pipelines in our experiments.
\section{Xtream}
\label{sec:method}

\cref{fig:overview} illustrates the main components of Xtream, a hybrid streaming reconstruction system comprising a 3DFM-based front-end that predicts relative transformations between sets of frames, and a back end that optimizes an explicit metric representation of the scene in the form of a factor graph. 
The input to Xtream is an image sequence $\{I_i\}^{N-1}_{i=0}$ containing $N$ frames and, optionally, synchronized accelerometer and gyroscope readings $\{(\bm{a}_j, \bm{\omega}_j)\}^{M-1}_{j=0}$ from an IMU rigidly attached to the camera with known extrinsics. Following established practice, we will assume that intrinsic and extrinsic sensor parameters are precalibrated in the visual-inertial setting. In the monocular setting, however, camera intrinsics may be either known or unknown.

\subsection{Front-end}
\label{sec:frontend}

\noindent\textbf{Stream Processing.} We partition the image stream into $K$ windows $\{\mathcal{W}_k\}^K_{k=1}$, each comprising $W$ frames $\{I_q^k\}^W_{q=1} \in \mathcal{W}_k$, with consecutive windows overlapping by $O$ frames, $\mathcal{O}_k = \mathcal{W}_{k-1} \cap \mathcal{W}_k$.
A frozen 3DFM jointly processes the frames in each window, predicting their corresponding up-to-scale camera poses ${\bar{{\bm{T}}}^k_q=\begin{pmatrix}{\mathtt{R}}_q^k &  {\bar{\mathbf{p}}}_q^k \\ \mathbf{0} & 1\end{pmatrix} \in SE(3)}$ and pixelwise depth maps $D^k_q$, both defined up to an unknown scale ${s}^k$, along with a depth confidence map $C^k_q$.
Additionally, we compute a DINOv2 SALAD~\cite{izquierdo2024optimal} descriptor $\mathbf{d}_i$ for each frame, which is used for graph densification and loop closure. 
When IMU measurements are available, we preintegrate them~\cite{forster2017onmanifold} over the interval $(t_{q}, t_{q+1}]$, using the current window's bias estimates $(\mathbf{b}_a^k, \mathbf{b}_\omega^k)$, to obtain rotation, translation and velocity residuals for the back end optimization.

\providecommand{\fade}[1]{\textcolor{black!45}{#1}}
\providecommand{\cfade}[1]{\textcolor{black!65}{#1}}   
\begin{table*}[!t]
\centering
\setlength{\aboverulesep}{0pt}\setlength{\belowrulesep}{0pt}\renewcommand{\arraystretch}{1.2}
\footnotesize
\setlength{\tabcolsep}{4pt}
\caption{\textbf{Absolute Trajectory Error on VBR.} Per block, best in \textbf{bold}, second best \underline{underlined}.}
\label{tab:ate_vbr}
\begin{threeparttable}
\begin{tabularx}{\textwidth}{@{}c@{\hspace{2pt}}l*{9}{>{\raggedleft\arraybackslash}X}}
\toprule
& Method & Colos0 & Campus0 & Campus1 & Pincio0 & Spagna0 & Diag0 & Ciamp1 & Avg. \\
\midrule
\multicolumn{9}{l}{\cellcolor{gray!12}\textbf{Monocular methods}, ATE RMSE [m] $\downarrow$ after Sim(3) alignment} &\cellcolor{gray!12} \\
\multirow{3}{*}{\rotatebox[origin=c]{90}{\scriptsize \cfade{Calibrated}}} & \cfade{{DROID-SLAM$^\ddagger$}} & \cfade{\underline{{{76.53}}}} & \cfade{{128.77}} & \cfade{\underline{{{63.89}}}} & \cfade{\textbf{{{43.86}}}} & \cfade{{50.63}} & \cfade{\textbf{{{35.58}}}} & \cfade{{132.84}} & \cfade{{76.01}} \\
 & \cfade{{DPV-SLAM$^\ddagger$}} & \cfade{{107.55}} & \cfade{\underline{{{56.85}}}} & \cfade{{88.46}} & \cfade{{102.64}} & \cfade{\textbf{{{32.63}}}} & \cfade{\underline{{{74.11}}}} & \cfade{\underline{{{60.95}}}} & \cfade{\underline{{{74.74}}}} \\
 & \cfade{{DPV-SLAM++$^\ddagger$}} & \cfade{\textbf{{{55.47}}}} & \cfade{\textbf{{{11.90}}}} & \cfade{\textbf{{{13.11}}}} & \cfade{\underline{{{101.84}}}} & \cfade{\underline{{{34.12}}}} & \cfade{{74.69}} & \cfade{\textbf{{{59.47}}}} & \cfade{\textbf{{{50.09}}}} \\
\cmidrule{2-10}
\multirow{10}{*}{\rotatebox[origin=c]{90}{\scriptsize Uncalibrated}} & VGGT-Long$^\ddagger$ & 45.73 & 132.68 & 115.65 & 64.40 & 58.49 & 33.66 & 187.97 & 91.23 \\
 & VGGT-SLAM 2.0$^\ddagger$ & 102.25 & 26.57 & 22.42 & 47.05 & 59.91 & 36.00 & 51.53 & 49.39 \\
 & DA3-Streaming & 45.47 & 99.18 & 8.87 & 24.44 & 59.07 & 35.40 & 36.16 & 44.09 \\
 & CUT3R$^\ddagger$ & 88.16 & 46.39 & 48.00 & 52.87 & 47.98 & 30.18 & 187.58 & 71.59 \\
 & TTT3R$^\ddagger$ & 81.63 & 68.36 & 65.67 & 35.56 & 39.03 & 19.84 & 185.59 & 70.81 \\
 & LoGeR$^\ddagger$ & 25.60 & 19.53 & 28.90 & 28.11 & 23.04 & 35.01 & 92.94 & 36.16 \\
 & CLoSeR$^\dagger$ & 14.23 & \underline{7.43} & \textbf{3.71} & \underline{9.78} & 13.04 & 12.92 & 16.93 & 11.15 \\
 & LingBot-Map$^\ddagger$ & 19.22 & 18.46 & 11.40 & 41.89 & 36.56 & 22.65 & 67.64 & 31.12 \\
 & AMB3R-SLAM$^\dagger$ & \textbf{3.95} & 9.29 & 6.73 & 10.75 & \underline{5.94} & \textbf{3.36} & \underline{11.92} & \underline{7.42} \\
 & Xtream (Ours) & \underline{6.80} & \textbf{6.24} & \underline{4.71} & \textbf{6.15} & \textbf{2.57} & \underline{3.62} & \textbf{7.74} & \textbf{5.40} \\
\midrule
\multicolumn{9}{l}{\cellcolor{gray!12}\textbf{Monocular-inertial methods}, ATE RMSE [m] $\downarrow$ after SE(3) alignment} &\cellcolor{gray!12} \\
& {ORB-SLAM3} & {$\times$} & {$\times$} & {$\times$} & {$\times$} & {$\times$} & {$\times$} & {$\times$} & {--} \\
& MASt3R-Fusion (global) & $\times$ & $\times$ & 30.72 & $\times$ & $\times$ & $\times$ & $\times$ & -- \\
& MASt3R-Fusion & 24.06 & 133.41 & 38.07 & 67.19 & 64.27 & 7.26 & 50.66 & 54.99 \\
& FoundationVIO & \underline{15.26} & \underline{9.67} & \underline{7.14} & \underline{3.34} & \underline{9.90} & \underline{5.09} & \underline{16.71} & \underline{9.59} \\
& Xtream (Ours) & \textbf{4.15} & \textbf{3.05} & \textbf{1.59} & \textbf{1.79} & \textbf{2.14} & \textbf{3.00} & \textbf{5.66} & \textbf{3.06} \\
\bottomrule
\end{tabularx}
\begin{tablenotes}
            \footnotesize
            \item[]$^\dagger$: from the method's own paper; $^\ddagger$: as reported by CLoSeR; the rest are run by us. $\times$: failed or diverged; --: not reported.
        \end{tablenotes}
        \end{threeparttable}
\end{table*}

\noindent\textbf{Scale consistency.}
For each window $\mathcal{W}_k$, we define an unknown logarithmic scale ratio $\ell_k \in \mathbb{R}^+$, such that multiplying the window's geometry (poses and depth maps) by $e^{\ell_k}$ converts it to metric scale. 
Consecutive windows are related through their first shared frame, $o_k \in \mathcal{O}_k$, for which both windows predict a depth map. The median depth ratio
\begin{equation}
  r_k = \operatorname*{median}_{\bm{u} \in \Omega} \frac{D^{k-1}_{o_k}(\bm{u})}{D^{k}_{o_k}(\bm{u})},
  \label{eq:ratio}
\end{equation}
computed over the image domain $\Omega$ and considering only pixels with valid depth predictions in both windows, converts the scale of $\mathcal{W}_k$ into that of $\mathcal{W}_{k-1}$.

\noindent\textbf{Keyframing.} From each window $\mathcal{W}_k$ we select a set $\mathcal{K}_k$ of $K$ keyframes from its newly observed frames $\mathcal{W}_k\setminus\mathcal{O}_k$, uniformly sampled in time and including the first and last new frames. The last selected frame also belongs to the next overlap $\mathcal{O}_{k+1}$, ensuring that consecutive windows remain connected through a shared keyframe.

\subsection{Back-end}
\label{sec:state}

Our backend is a factor graph~\cite{dellaert2021factor} $\mathcal{G}=(\mathcal{V}, \mathcal{F}, \mathcal{E})$, comprising a set of nodes $\mathcal{V}=\{\bm{n}_i, \gamma^{\mathrm{imu}}\bm{b}_{\beta}\}$ representing the variables to be estimated, a set of factors $\mathcal{F} =\{ \bm{e}^{\mathrm{v}}_{i},\gamma^{\mathrm{imu}}\bm{e}^{\mathrm{b}}_{\beta},\gamma^{\mathrm{imu}}\bm{e}^{\mathrm{imu}}_{i}e^{\mathrm{s}}_k,e^{\mathrm{s},\mathcal{L}}_{a,b},\gamma^{\mathrm{m}}e^{\mathrm{m}}_k\}$ encoding the constraints between them, and a set of edges $\mathcal{E}$ connecting variables to their corresponding factors. The binary indicators $\gamma^{\mathrm{imu}}$ and $\gamma^{\mathrm{m}}$ determine whether metric-scale measurements are provided by IMU readings or 3DFMs, respectively. Each node $\bm{n}_i = (\bm{T}_i, \gamma^{\mathrm{imu}}\bm{v}_i) \in SE(3)\times\mathbb{R}^3 $ contains the camera pose of the $i$-th frame in a gravity-aligned world frame, and its velocity, respectively. The graph also includes an IMU bias $\bm{b}_\beta \in \mathbb{R}^6$ for every $R$ consecutive windows, as well as a logarithmic scale ratio for every window $\ell_k$ and for every loop $\ell^\mathcal{L}_{i,j}$.
These scale ratios are indirectly constrained either by inertial factors or by metric-aware 3DFMs, with their estimates propagated throughout the reconstruction through the factor graph.

\noindent\textbf{Visual factors.}
We introduce relative pose constraints between keyframe pairs through the up-to-scale 3DFM pose predictions ${\Delta{\bar{\bm{T}}}^k_{i,i+1} = ({\bar{\bm{T}}}^k_{i})^{-1}\bar{{\bm{T}}}^k_{i+1}}$. These measurements are related to the corresponding relative poses in the metric state $ {\bm{T}_{i,i+1} = \bm{T}_i^{-1}\bm{T}_{i+1}}$ through the following residual:
\begin{equation}
  \bm{e}^{\mathrm{v}}_{i} = \operatorname{Log}\!\left( \Delta{\bar{\bm{T}}}^k_{i,i+1} \begin{pmatrix}{\mathtt{R}}_{i,i+1} &  e^{-\ell_k}{\mathbf{p}}_{i,i+1} \\ \mathbf{0} & 1\end{pmatrix}\, \right) \in \mathbb{R}^6,
  \label{eq:vision_factor}
\end{equation}
with diagonal noise $(\sigma_R\,\bm{1}_3,\; \sigma_t e^{-\ell_k}\bm{1}_3)$. Note that the first frame in each window corresponds to an overlapping frame from the previous one, which anchors the two windows. Visual residuals are expressed in 3DFM units, making them invariant to changes in global scale and preventing scale collapse. 

\noindent\textbf{Inertial factors.}
Inertial measurements introduce two types of residuals: a relative pose constraint between consecutive frames and a bias constraint.
The first residual $\bm{e}^{\mathrm{imu}}_{i}$ relates the preintegrated IMU measurements to the states $(\bm{n}_{i}, \bm{n}_{i+1}, \bm{b}_{\beta})$~\cite{forster2017onmanifold}. This residual provides metric-scale and gravity constraints on the state.
The bias residual is imposed between consecutive blocks of $R$ windows $(\beta, \beta+1)$ as:
\begin{equation}
  \bm{e}^{\mathrm{b}}_{\beta} = \bm{b}_{\beta+1} - \bm{b}_{\beta} \sim \mathcal{N}\big(\bm{0},\, T_\beta\, \mathrm{diag}(\sigma^2_{wa}\bm{1}_3, \sigma^2_{wg}\bm{1}_3)\big),
  \label{eq:bias_factor}
\end{equation}
with $T_\beta$ the duration of the current block, and $(\sigma^2_{wa}, \sigma^2_{wg})$ the random walk variances. 

\noindent\textbf{Scale factors.}
Following~\cref{eq:ratio}, the relative scale between consecutive windows is constrained through the following residual:
\begin{equation}
  e^{\mathrm{s}}_k = \ell_k - \ell_{k-1} - \log r_k,
  \label{eq:scale_factor}
\end{equation}
with standard deviation $\sigma_{\mathrm{s}}$.

\paragraph{Mid- and long-term loops.}
The Xtream front-end only establishes constraints between consecutive windows through their overlap, without explicitly accounting for longer-range scene structure.
We introduce loops between co-visible windows separated by a few windows, thereby improving local consistency, as well as between temporally and spatially distant windows, providing global consistency through loop closures. Each loop is one 3DFM feedforward pass over
frames of both windows, $\mathcal{L}_{i,j}$.

\noindent\emph{Graph densification.}
For each keyframe in a new window $\mathcal{W}_j$, we search the keyframes of the previous three windows for DINOv2 SALAD~\cite{izquierdo2024optimal}
cosine similarity over a threshold $\delta$, excluding the overlapping frames. Each previous window $\mathcal{W}_i$ with a potential match yields one covisibility loop,
and the union of both windows' keyframes, $\mathcal{L}^{\mathrm{cov}}_{i,j} = \mathcal{K}_i \cup \mathcal{K}_j$ is processed by the 3DFM for relative pose predictions.

\noindent\emph{Closures.}
Following CLoSeR streaming detector~\cite{li2026closer}, Xtream first identifies loop closure candidates based on SALAD similarity scores over $\delta$, separated by at least 4 windows. A geometric filter is then applied to the detected candidates, retaining those with a chord-to-arc ratio below $0.85$ along the chained trajectory, followed by non-maximum suppression over two windows. Each resulting loop $\mathcal{L}^{\mathrm{lc}}_{i,j}$ combines the matched frames with up to $(W{-}2O)/2$ adjacent frames on each side. 

\noindent\textbf{Loop factors.}
From the 3DFM relative poses, we define a factor $\bm{e}^{\mathcal{L}}_{i,j}$, of the same form as~\eqref{eq:vision_factor}, for both loop types: for the middle frames $a$ and $b$ of each side of $\mathcal{L}_{i,j}$, the 3DFM up-to-scale relative pose $\Delta\bar{\bm{T}}^{\mathcal{L}}_{a,b}$ constrains their metric poses $\bm{T}_{a}$ and $\bm{T}_{b}$. Each loop has its own logarithmic scale ratio $\ell^{\mathcal{L}}_{i,j}$, a Huber loss and noise $(\sigma^{\mathcal{L}}_R,\ \sigma^{\mathcal{L}}_t e^{-\hat\ell^{\mathcal{L}}_{i,j}})/\sqrt{w_{i,j}}$, with $w_{i,j}$ a weight described later.

\noindent\textbf{Loop scale factors.}
A loop relates the middle frames from two windows, $\mathcal{W}_i$ and $\mathcal{W}_j$, each with its own logarithmic scale ratio $\ell_i$ and $\ell_j$. However, as the loop window $\mathcal{L}_{i,j}$ has also been jointly processed by the 3DFM, it will also have its own logarithmic scale ratio $\ell^{\mathcal{L}}_{i,j}$. Rather than directly constraining $\ell_j - \ell_i$, a loop acts through its own logarithmic scale ratio $\ell^{\mathcal{L}}_{i,j}$, related to each window's ratio by a separate factor, and passing in this manner information between the two. For each of the two windows $c \in \{i, j\}$ that are related by a loop, we apply Umeyama to align the window's own 3DFM poses of $\mathcal{S}_c$, the loop's frames from that window (a fixed measurement in that window's unit), onto the loop pass's poses of the same frames. This produces an estimated relative scale $s^{\mathcal{L}}_{c}$, in loop units per window-$c$ unit, which constrains the two ratios:
\begin{equation}
  e^{\mathrm{s},\mathcal{L}}_{c} = \ell_c - \ell^{\mathcal{L}}_{i,j} - \log s^{\mathcal{L}}_{c}, \qquad c \in \{i, j\},
  \label{eq:loop_scale_factor}
\end{equation}
with standard deviation $\big((\sigma^{\mathcal{L}}_s)^2 + \sigma^2_{s,c}\big)^{1/2}/\sqrt{w_{i,j}}$, where $\sigma_{s,c}$ is the standard error of the aligned log scale; which grows if the frames of $\mathcal{S}_c$ barely move, so a side without baseline gets a loose factor. $\ell^{\mathcal{L}}_{i,j}$ is initialised from the more certain side. Eliminating $\ell^{\mathcal{L}}_{i,j}$ recovers the direct relation, $\ell_j - \ell_i \approx \log s^{\mathcal{L}}_j - \log s^{\mathcal{L}}_i$, now weighted per side.

\noindent\textbf{Loop verification and weighting.}
A wrong loop, either from perceptual aliasing or a failed loop pass, can corrupt the whole trajectory. Each loop is therefore checked or weighted by its agreement with the current state.
In the monocular scenario, we follow CLoSeR's approach: after each solve, loops whose translation residual, measured in the loop's own scale, exceeds a threshold are removed and the graph is re-solved. Conversely, in the inertial approach, the IMU serves as an additional source to check for disagreement, and loops are weighted \emph{before} they enter the graph. The innovation of $\bm{e}^{\mathcal{L}}_{i,j}$, with the 3DFM relative pose scaled, has a covariance that combines the graph's joint marginal of $(\bm{T}_a, \bm{T}_b, \ell)$ with the loop's own noise.
A loop's unit is uncertain by construction, so we minimise the Mahalanobis distance over the loop's log scale. This leaves $d^2$ on the $p = 5$ scale-free components, so that only the loop's shape is judged.
Innovations are heavy-tailed, with a scale $\lambda$ that reflects the overconfidence of the window vision factors. We take $\lambda$ from the graph itself, as the mean whitened squared residual of its window vision factors times a constant $\kappa$, and apply the Student-$t$ weight, with $\nu$ degrees of freedom
\begin{equation}
  w_{i,j} = \min\!\Big(1,\; \frac{\nu + p}{\nu + d^2 / \lambda}\Big).
  \label{eq:weight}
\end{equation}
With an IMU, a loop is never removed after the fact, and the graph never re-solves to undo it.

\noindent\textbf{Metric scale prior from a 3DFM.}
Handovers and covisibility edges constrain only differences $\ell_j - \ell_i$ of the window logarithmic scale ratios, so in the
monocular setting a small systematic error per handover accumulates over hundreds of windows.
We propose adding an absolute measurement for each $\ell_k$ through the metric predictions of a 3DFM.

\noindent\emph{Measurement.}
Let $D^k_f : \Omega \to \mathbb{R}_{>0}$ be the depth map of frame $f$ predicted by the pass of window
$\mathcal{W}_k$, in that window's unit, and $D^{\mathrm{m}}_f : \Omega \to \mathbb{R}_{>0}$ the depth map of the
same frame predicted in metres by a monocular metric-depth model (\emph{e.g.}, \emph{DA3-Metric-Large}).
Over the keyframes $\mathcal{K}_k \subset \mathcal{W}_k$, the window's metres per unit are measured as
\begin{equation}
  m_k \;=\; \operatorname*{med}_{f \in \mathcal{K}_k}\;
  \frac{\operatorname{med}_{u \in \Omega} D^{\mathrm{m}}_f(u)}{\operatorname{med}_{u \in \Omega} D^{k}_f(u)} ,
  \label{eq:metre}
\end{equation}
where the medians make $m_k$ robust to outlier scale predictions and to a minority of failed keyframes.
Since $D^k_f$ and the translations of $\mathcal{W}_k$ share one unit, $m_k$ also converts the window's
translations to metres, \emph{i.e.}, it measures $e^{\ell_k}$.
Each $m_k$ depends only on the images of $\mathcal{W}_k$, so its error does not accumulate along the sequence.

\noindent\emph{Factor.}
Xtream's robust unary factor on the window's logarithmic scale ratio is as follows,
\begin{equation}
  e^{\mathrm{m}}_k = \rho\!\left( \frac{\ell_k - \log m_k}{\sigma_k} \right),
  \label{eq:metric_prior}
\end{equation}
with $\rho$ the Huber loss 
and $\ell_0$ initialised at $\log m_0$.

As the link already relates consecutive scales, the residual $\varepsilon_k = \log m_k - \log m_{k-1} - \log r_k$ measures the metric estimate's error, and $\sigma_k$ is $3\times$ the robust standard deviation ($1.4826\,\mathrm{MAD}$) of the residuals seen so far, fixed when the factor is inserted.
The prior is only added when the window moves: without baseline, its depth scale does not determine its pose scale.
We require the camera centres $\bm{c}_f$ of $\mathcal{W}_k$ to spread by at least 1\% of the window's median depth.
With an IMU the prior is not used: the inertial factors make $\ell_k$ observable, more accurately than the
per-window metric estimate.

\subsection{Initialisation and online processing}
\label{sec:solver}

\noindent\textbf{Monocular setting.}
The first scale is initialised
from the metric prior, $\ell_0 = \log m_0$. Graph factors constrain only \emph{relative} scale, so a small systematic error per handover accumulates over hundreds of windows. The per-window metric prior adds the \emph{absolute} scale measurement that bounds it. 

\noindent\textbf{Monocular-inertial setting.}
When an IMU is available, the initial pose orientation is instead initialized by transforming the accelerometer readings across the initial temporal window into the first frame and aligning their mean with the upward vertical direction.
The first window scale factor $\ell_0$ is seeded by the GT-free visual-inertial initialization of VI3~\cite{lozano2026vi3}.
The seed enters as a prior on $\ell_0$. Later windows are initialized by chaining~\eqref{eq:scale_factor}.

New graph poses are chained from the window's relative poses at its current scale; velocities start from finite differences and biases from the previous estimate.

\noindent\textbf{Pruning.} Windows are grouped into blocks of $\mathcal{B}$ consecutive windows. One of every two consecutive blocks is pruned. In each of its windows $\mathcal{W}_k$, only the keyframes $\mathcal{K}_k$ and frames in loop factors remain. Between consecutive remaining frames, visual factors are grouped, with their noise propagated to first order. For the mono-inertial setting, inertial measurements are preintegrated again into a single factor at the block's bias estimate. A frame pruned from the factor graph keeps a relative pose as estimated at pruning and receives interpolated fractions of the corrections of close keyframes in the graph.

\noindent\textbf{Solver.}
The graph is solved by Levenberg--Marquardt, warm-started from the current estimate. Each step relinearizes the full graph, which keyframe pruning keeps small. A window update adds its factors and takes a single step. A loop update iterates to convergence, and so does a final solve when the stream ends.


\subsection{Metric dense reconstruction}
\label{sec:map}

The explicit state turns the model's depth maps into a metric, globally consistent map at no extra estimation cost. Each frame $f$ is taken from the first window $k(f)$ that predicts it, back-projected with the window's predicted intrinsics into a pointmap $\bm{P}^{k(f)}_f$, scaled by that window's metre, and placed at the state's pose:

\begin{align}
  \mathcal{M} =& \bigcup_f \Big\{ \bm{X}_f \big( e^{\ell_{k(f)}} \bm{P}^{k(f)}_f(\bm{p}) \big) \;:\;\\
  &C^{k(f)}_f(\bm{p}) \ge c_{k(f)},\; e^{\ell_{k(f)}} \lVert \bm{P}^{k(f)}_f(\bm{p}) \rVert \le \rho \Big\},
  \label{eq:map}
\end{align}
where $c_k$ is a per-window confidence percentile and $\rho$ a maximum range. For extra accuracy, the map frames can be re-predicted (additional pass) with the state as input: groups of consecutive frames are passed through the 3DFM conditioned on their poses $\bm{T}_f$, so the depth is predicted for the state's poses rather
than scaled after the fact, and the pointmaps in Eq.~\eqref{eq:map} are replaced by the conditioned ones.

\begin{figure}[!h]
  \centering
  \includegraphics[width=.9\columnwidth]{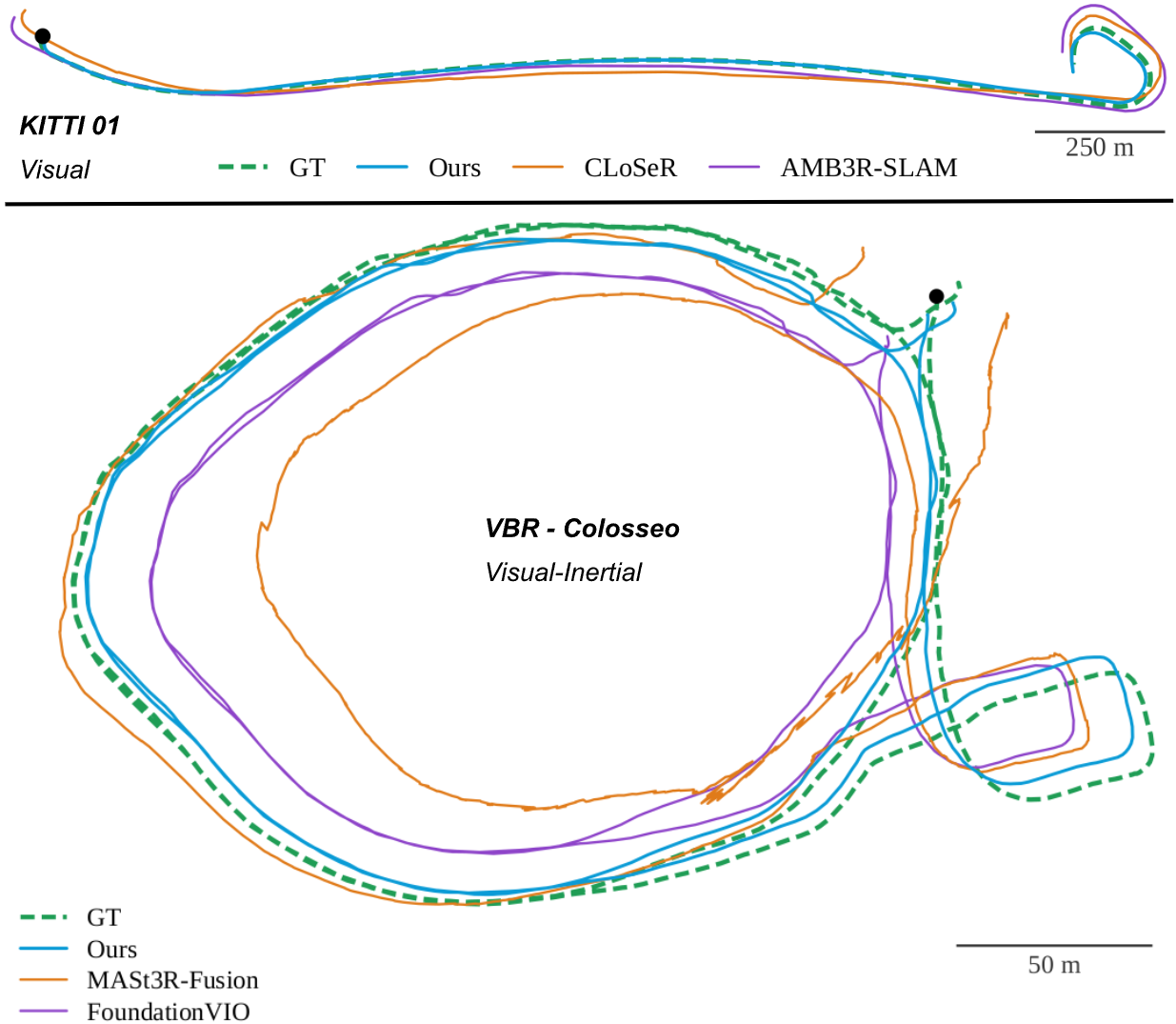}
  \caption{\textbf{Qualitative comparison.} Top view of full trajectories against GT and baselines, aligned as for the ATE.}
  \label{fig:qualitative}
\end{figure}
\section{Experiments}
\label{sec:experiments}

\providecommand{\fade}[1]{\textcolor{black!45}{#1}}
\providecommand{\cfade}[1]{\textcolor{black!65}{#1}}   
\begin{table*}[h!]
\centering
\setlength{\aboverulesep}{0pt}\setlength{\belowrulesep}
{0pt}\renewcommand{\arraystretch}{1.2}
\footnotesize
\caption{\textbf{Absolute trajectory error on Oxford Spires.} Per block, best in \textbf{bold}, second best \underline{underlined}.}
\label{tab:ate_oxford_spires}
\begin{threeparttable}
\resizebox{\textwidth}{!}{
\begin{tabular}{@{}c@{\hspace{2pt}}lrrrrrrrrrrrr}
\toprule
& Method & Keb02 & Keb03 & Keb04 & Keb05 & Obs01 & Obs02 & Ble05 & CC02 & CC03 & CC05 & Bod02 & Avg. \\
\midrule
\multicolumn{13}{l}{\cellcolor{gray!12}\textbf{Monocular methods}, ATE RMSE [m] $\downarrow$ after Sim(3) alignment} & \cellcolor{gray!12}\\
\multirow{3}{*}{\rotatebox[origin=c]{90}{\scriptsize \cfade{Calibrated}}} & \cfade{{DROID-SLAM$^\ddagger$}} & \cfade{{6.66}} & \cfade{{29.71}} & \cfade{\textbf{{{14.10}}}} & \cfade{{36.60}} & \cfade{\textbf{{0.17}}} & \cfade{\textbf{{0.36}}} & \cfade{{15.88}} & \cfade{{10.17}} & \cfade{\underline{{{0.33}}}} & \cfade{\textbf{{{8.92}}}} & \cfade{{5.64}} & \cfade{{11.69}} \\
 & \cfade{{DPV-SLAM$^\ddagger$}} & \cfade{\underline{{{5.84}}}} & \cfade{\textbf{{{4.53}}}} & \cfade{{19.74}} & \cfade{\underline{{{25.25}}}} & \cfade{{0.27}} & \cfade{{6.24}} & \cfade{\textbf{{{11.62}}}} & \cfade{\textbf{{1.15}}} & \cfade{{0.94}} & \cfade{{42.62}} & \cfade{\underline{{3.51}}} & \cfade{\underline{{{11.06}}}} \\
 & \cfade{{DPV-SLAM++$^\ddagger$}} & \cfade{\textbf{{{1.33}}}} & \cfade{\underline{{{29.35}}}} & \cfade{\underline{{{18.39}}}} & \cfade{\textbf{{{3.66}}}} & \cfade{\underline{{0.20}}} & \cfade{\underline{{{2.94}}}} & \cfade{\underline{{{12.06}}}} & \cfade{\underline{{{7.82}}}} & \cfade{\textbf{{0.28}}} & \cfade{\underline{{{40.19}}}} & \cfade{\textbf{{3.32}}} & \cfade{\textbf{{{10.87}}}} \\
\cmidrule{2-14}
\multirow{10}{*}{\rotatebox[origin=c]{90}{\scriptsize Uncalibrated}} & VGGT-Long$^\ddagger$ & 3.63 & 14.00 & 32.66 & 4.64 & 3.80 & 5.64 & 11.41 & 9.03 & 4.02 & 14.51 & 16.76 & 10.92 \\
 & VGGT-SLAM 2.0$^\ddagger$ & 7.28 & 20.57 & 6.90 & 17.06 & 1.76 & 1.85 & \textbf{1.26} & \textbf{1.64} & 0.96 & 16.41 & 8.51 & 7.65 \\
 & DA3-Streaming & \underline{1.04} & 1.10 & 4.54 & 1.17 & \underline{{1.37}} & \underline{{1.06}} & 6.33 & 8.66 & 1.16 & 3.63 & 7.29 & 3.40 \\
 & CUT3R$^\ddagger$ & 25.44 & 17.75 & 42.51 & 36.12 & 26.83 & 27.05 & 49.45 & 41.67 & 11.83 & 32.47 & 56.71 & 33.44 \\
 & TTT3R$^\ddagger$ & 29.05 & 17.45 & 32.54 & 41.37 & 26.37 & 26.65 & 35.25 & 41.46 & 14.18 & 27.26 & 69.42 & 32.82 \\
 & LoGeR$^\ddagger$ & 4.01 & 3.11 & 6.51 & 5.76 & 7.01 & 6.98 & 6.98 & 6.72 & 1.84 & 10.22 & 14.09 & 6.66 \\
 & CLoSeR$^\dagger$ & 2.75 & 2.40 & 5.79 & 4.65 & 4.33 & 4.03 & 5.06 & 5.70 & 1.19 & 8.03 & 14.08 & 5.27 \\
 & LingBot-Map$^\ddagger$ & 3.42 & 3.27 & 3.00 & 9.39 & 2.81 & 2.50 & 2.78 & 13.39 & 2.94 & 28.79 & 9.27 & 7.41 \\
 & AMB3R-SLAM$^\dagger$ & \textbf{0.80} & \textbf{0.55} & \underline{2.06} & \underline{1.15} & 1.78 & 1.74 & 6.74 & \underline{{2.23}} & \textbf{0.25} & \textbf{2.09} & \underline{6.11} & \underline{2.32} \\
 & Xtream (Ours) & 1.46 & \underline{0.72} & \textbf{1.26} & \textbf{1.05} & \textbf{{0.87}} & \textbf{0.87} & \underline{1.33} & 3.34 & \underline{0.28} & \underline{2.17} & \textbf{{5.00}} & \textbf{1.67} \\
\midrule
\multicolumn{13}{l}{\cellcolor{gray!12}\textbf{Monocular-inertial methods}, ATE RMSE [m] $\downarrow$ after SE(3) alignment} &\cellcolor{gray!12} \\
& {ORB-SLAM3} & {0.88} & \textbf{0.53} & \underline{0.70} & \underline{1.24} & \underline{0.74} & {4.38} & \textbf{0.93} & \underline{0.43} & {0.20} & \textbf{0.54} & {5.52} & \underline{1.46} \\
& MASt3R-Fusion (global) & 2.28 & {2.21} & 4.29 & {1.78} & 5.55 & 6.05 & $\times$ & \textbf{0.42} & \textbf{0.15} & {0.79} & \textbf{3.57} & -- \\
& MASt3R-Fusion & 3.07 & 2.21 & 5.47 & 6.10 & 5.76 & 6.15 & 2.55 & 2.62 & 0.34 & 2.12 & \underline{3.59} & 3.64 \\
& FoundationVIO & \underline{0.85} & 3.72 & {2.33} & 2.93 & {1.32} & \underline{1.01} & \underline{2.15} & 1.05$^\dagger$ & \underline{0.16$^\dagger$} & 1.92$^\dagger$ & 9.26 & {2.43} \\
& Xtream (Ours) & \textbf{0.69} & \underline{0.70} & \textbf{0.58} & \textbf{0.64} & \textbf{0.33} & \textbf{0.56} & {2.62} & {0.61} & 0.63 & \underline{0.73} & 4.10 & \textbf{1.11} \\
\bottomrule
\end{tabular}}
\begin{tablenotes}
            \footnotesize
            \item[]$^\dagger$: from the method's own paper; $^\ddagger$: as reported by CLoSeR; the rest are run by us. $\times$: failed or diverged; --: not reported.
        \end{tablenotes}
        \end{threeparttable}
\end{table*}

\providecommand{\fade}[1]{\textcolor{black!45}{#1}}
\providecommand{\cfade}[1]{\textcolor{black!65}{#1}}   
\begin{table*}[h!]
\centering
\setlength{\aboverulesep}{0pt}\setlength{\belowrulesep}{0pt}\renewcommand{\arraystretch}{1.2}
\footnotesize
\setlength{\tabcolsep}{4pt}
\caption{\textbf{Absolute Trajectory Error on KITTI.} Per block, best in \textbf{bold}, second best \underline{underlined}.}
\label{tab:ate_kitti}
\begin{threeparttable}
\begin{tabularx}{\textwidth}{@{}c@{\hspace{2pt}}l*{12}{>{\raggedleft\arraybackslash}X}}
\toprule
& Method & 00 & 01 & 02 & 03 & 04 & 05 & 06 & 07 & 08 & 09 & 10 & Avg. \\
\midrule
\multicolumn{13}{l}{\cellcolor{gray!12}\textbf{Monocular methods}, ATE RMSE [m] $\downarrow$ after Sim(3) alignment} & \cellcolor{gray!12} \\
\multirow{3}{*}{\rotatebox[origin=c]{90}{\scriptsize \cfade{Calibrated}}} & \cfade{{DROID-SLAM$^\ddagger$}} & \cfade{\underline{92.10}} & \cfade{344.60} & \cfade{\underline{107.61}} & \cfade{\textbf{2.38}} & \cfade{1.00} & \cfade{118.50} & \cfade{62.47} & \cfade{21.78} & \cfade{161.60} & \cfade{\textbf{72.32}} & \cfade{118.70} & \cfade{100.28} \\
 & \cfade{{DPV-SLAM$^\ddagger$}} & \cfade{112.80} & \cfade{\textbf{11.50}} & \cfade{123.53} & \cfade{\underline{2.50}} & \cfade{\underline{0.81}} & \cfade{\underline{57.80}} & \cfade{\underline{54.86}} & \cfade{\underline{18.77}} & \cfade{\textbf{110.49}} & \cfade{\underline{76.66}} & \cfade{\textbf{13.65}} & \cfade{\underline{53.03}} \\
 & \cfade{{DPV-SLAM++$^\ddagger$}} & \cfade{\textbf{8.30}} & \cfade{\underline{11.86}} & \cfade{\textbf{39.64}} & \cfade{\underline{2.50}} & \cfade{\textbf{0.78}} & \cfade{\textbf{5.74}} & \cfade{\textbf{11.60}} & \cfade{\textbf{1.52}} & \cfade{\underline{110.90}} & \cfade{76.70} & \cfade{\underline{13.70}} & \cfade{\textbf{25.75}} \\
\cmidrule{2-14}
\multirow{10}{*}{\rotatebox[origin=c]{90}{\scriptsize Uncalibrated}} 
 & VGGT-Long$^\ddagger$ & 8.67 & 121.17 & 32.08 & 6.12 & 4.23 & 8.31 & 5.34 & 4.63 & 53.10 & 41.99 & 18.37 & 27.64 \\
 & VGGT-SLAM 2.0$^\ddagger$ & 108.39 & 155.00 & 140.15 & 13.45 & 2.87 & 10.12 & 29.17 & 17.27 & 75.33 & 27.70 & 24.09 & 54.87 \\
 & DA3-Streaming & 8.98 & 193.09 & 50.09 & 2.90 & 1.46 & 4.17 & \underline{3.17} & 3.70 & 38.23 & 9.22 & 11.91 & 29.72 \\
 & CUT3R$^\ddagger$ & 190.38 & 90.59 & 264.39 & 20.40 & 7.31 & 92.25 & 67.54 & 22.48 & 145.08 & 67.42 & 40.00 & 91.62 \\
 & TTT3R$^\ddagger$ & 119.94 & 99.59 & 238.07 & 16.83 & 3.98 & 36.38 & 47.20 & 11.62 & 107.33 & 86.96 & 33.58 & 72.86 \\
 & LoGeR$^\ddagger$ & 62.34 & \underline{41.64} & 39.64 & 4.89 & 1.82 & 41.27 & 13.99 & 16.24 & 26.46 & 22.71 & 8.84 & 25.44 \\
 & CLoSeR$^\dagger$ & 6.58 & 41.80 & \textbf{28.38} & 4.88 & 1.82 & 6.58 & 4.77 & 4.87 & 26.51 & \underline{5.64} & 8.83 & \underline{12.79} \\
 & LingBot-Map$^\ddagger$ & 23.22 & 88.84 & 71.15 & \underline{2.81} & \underline{0.91} & 17.40 & 4.91 & 5.16 & \textbf{14.24} & 21.71 & 14.54 & 24.08 \\
 & AMB3R-SLAM$^\dagger$ & \textbf{3.67} & 55.91 & \underline{31.55} & 4.80 & 2.04 & \underline{2.68} & 3.33 & \underline{1.73} & \underline{23.56} & 9.03 & \textbf{5.96} & 13.11 \\
 & Xtream (Ours) & \underline{4.15} & \textbf{8.13} & 51.84 & \textbf{0.93} & \textbf{0.44} & \textbf{2.01} & \textbf{2.27} & \textbf{1.52} & 24.27 & \textbf{3.76} & \underline{8.37} & \textbf{9.79} \\
\bottomrule
\end{tabularx}
\begin{tablenotes}
            \footnotesize
            \item[]$^\dagger$: from the method's own paper; $^\ddagger$: as reported by CLoSeR; the rest are run by us.
        \end{tablenotes}
        \end{threeparttable}
\end{table*}

\subsection{Comparisons against baselines}

Table \ref{tab:ate_vbr} reports our results on VBR. We substantially reduce the ATE \cite{grupp2017evo} compared with calibrated and uncalibrated monocular baselines, as well as monocular-inertial methods. Notably, CLoSeR and AMB3R-SLAM already achieve substantial improvements over previous approaches, thanks to their contributions in graph densification and scale consistency. Our explicit metric-scale model brings further gains over these methods, establishing a new state of the art. \cref{fig:qualitative} shows a qualitative comparison against the best-performing monocular and mono-inertial baselines in two sequences from VBR and KITTI.

Tables \ref{tab:ate_oxford_spires} and \ref{tab:ate_kitti} report our results on Oxford Spires and KITTI, respectively. Again, we substantially reduce the ATE compared to both calibrated and uncalibrated monocular baselines, as well as published and concurrent monocular-inertial baselines. The results on KITTI follow a similar trend to those on VBR: Xtream outperforms both CLoSeR and AMB3R-SLAM, and all three methods substantially beat the remaining baselines. Trajectories are generally shorter in Oxford Spires, resulting in less accumulated drift, particularly scale drift. Consequently, ATE values are generally lower and the performance gap between methods is narrower. Nevertheless, our method still achieves a significant reduction in ATE over the second-best baseline, the concurrent work AMB3R-SLAM.

Tables \ref{tab:ate_euroc} and \ref{tab:ate_scarf} report our results on the EuRoC and ScaRF-SLAM datasets, respectively. These datasets cover substantially smaller environments than previous benchmarks, with trajectories spanning only tens of meters, resulting in generally lower ATE values. Since scale drift has a more limited impact at these spatial scales, Xtream achieves performance comparable to the concurrent work AMB3R-SLAM in the monocular setting. Both methods substantially outperform CLoSeR, which we attribute to their denser graph models, incorporating factors between non-consecutive keyframes. In the monocular-inertial setting, ORB-SLAM3 achieves the best results on EuRoC, benefiting from its highly optimized, feature-based pipeline, fitting the specifications of this benchmark. Xtream, however, achieves the best performance among 3DFM-based methods. On ScaRF-SLAM, Xtream once again establishes the state of the art.

\providecommand{\fade}[1]{\textcolor{black!45}{#1}}
\begin{table}[h!]
\centering
\setlength{\aboverulesep}{0pt}\setlength{\belowrulesep}{0pt}\renewcommand{\arraystretch}{1.2}
\footnotesize
\setlength{\tabcolsep}{2pt}
\caption{\textbf{Absolute Trajectory Error on EuRoC.} Per block, best in \textbf{bold}, second best \underline{underlined}.}
\label{tab:ate_euroc}
\resizebox{\columnwidth}{!}{
\begin{tabular}{l*{12}{r}}
\toprule
Method & MH1 & MH2 & MH3 & MH4 & MH5 & V11 & V12 & V13 & V21 & V22 & V23 & Avg. \\
\midrule
\multicolumn{12}{l}{\cellcolor{gray!12}\textbf{Monocular methods}, ATE RMSE [m] $\downarrow$ after Sim(3) alignment} & \cellcolor{gray!12} \\
DA3-Streaming & 0.37 & 0.34 & 0.14 & \textbf{0.17} & 0.31 & 0.07 & 0.10 & \underline{0.08} & 0.08 & 0.06 & 0.13 & 0.17 \\
LoGeR$^\ddagger$ & 0.73 & 0.88 & 1.43 & 2.17 & 2.53 & 0.34 & 0.47 & 0.54 & 0.28 & 0.50 & 0.83 & 0.97 \\
CLoSeR$^\dagger$ & 0.40 & 0.48 & 0.52 & 0.77 & 0.91 & 0.12 & 0.19 & 0.29 & 0.19 & 0.24 & 0.30 & 0.40 \\
AMB3R-SLAM & \underline{0.06} & \underline{0.07} & \textbf{0.06} & 0.26 & \underline{0.12} & \textbf{0.03} & \textbf{0.05} & \textbf{0.07} & \underline{0.04} & \underline{0.05} & \underline{0.10} & \underline{0.08} \\
Xtream (Ours) & \textbf{0.04} & \textbf{0.06} & \underline{0.06} & \underline{0.18} & \textbf{0.09} & \underline{0.04} & \underline{0.05} & 0.09 & \textbf{0.03} & \textbf{0.03} & \textbf{0.08} & \textbf{0.07} \\
\midrule
\multicolumn{12}{l}{\cellcolor{gray!12}\textbf{Monocular-inertial methods}, ATE RMSE [m] $\downarrow$ after SE(3) alignment} & \cellcolor{gray!12} \\
ORB-SLAM3$^\dagger$ & \textbf{0.06} & \textbf{0.04} & \textbf{0.05} & \textbf{0.07} & \textbf{0.06} & \textbf{0.05} & \textbf{0.01} & \textbf{0.04} & \textbf{0.04} & \textbf{0.02} & \textbf{0.03} & \textbf{0.04} \\
MASt3R-Fusion (global) & 0.33 & $\times$ & 0.09 & 0.22 & \underline{0.18} & 0.07 & \underline{0.03} & \textbf{0.04} & 0.16 & \underline{0.04} & \underline{0.04} & -- \\
MASt3R-Fusion & 0.37 & 0.67 & 0.44 & 0.48 & 0.54 & 0.14 & 0.12 & \underline{0.09} & 0.27 & 0.13 & 0.31 & 0.32 \\
FoundationVIO & \underline{0.27} & 0.27 & 0.28 & 0.68 & 0.74 & 0.15 & 0.28 & 0.30 & 1.84 & 0.67 & 0.20 & 0.52 \\
Xtream (Ours) & \textbf{0.06} & \underline{0.13} & \underline{0.07} & \underline{0.17} & 0.19 & \underline{0.06} & 0.07 & \underline{0.09} & \underline{0.11} & 0.10 & 0.14 & \underline{0.11} \\
\bottomrule
\end{tabular}}
\end{table}

\providecommand{\fade}[1]{\textcolor{black!45}{#1}}
\begin{table}[h!]
\centering
\setlength{\aboverulesep}{0pt}\setlength{\belowrulesep}{0pt}\renewcommand{\arraystretch}{1.2}
\footnotesize
\setlength{\tabcolsep}{3pt}
\caption{\textbf{Absolute Trajectory Error on ScaRF-SLAM.} Per block, best in \textbf{bold}, second best \underline{underlined}.}
\label{tab:ate_scarf}
\ifdefined\scarfhdrw\else\newlength{\scarfhdrw}\fi
\settowidth{\scarfhdrw}{\textbf{Monocular-inertial methods}, ATE RMSE [m] $\downarrow$ after SE(3) alignment}
\resizebox{\columnwidth}{!}{
\begin{tabular*}{\dimexpr\scarfhdrw+2\tabcolsep\relax}{@{\extracolsep{\fill}}l*{6}{r}}
\toprule
Method & r01 & r02 & r03 & r04 & r05 & Avg. \\
\midrule
\multicolumn{7}{l}{\cellcolor{gray!12}\textbf{Monocular methods}, ATE RMSE [m] $\downarrow$ after Sim(3) alignment} \\
DA3-Streaming & 0.17 & 0.07 & \underline{0.11} & 0.16 & 0.20 & 0.14 \\
LoGeR$^\ddagger$ & 1.26 & 0.77 & 0.58 & 0.75 & 0.90 & 0.85 \\
CLoSeR$^\dagger$ & 0.52 & 0.26 & 0.23 & 0.32 & 0.13 & 0.29 \\
AMB3R-SLAM & \textbf{0.13} & \textbf{0.04} & \textbf{0.05} & \underline{0.09} & \textbf{0.04} & \textbf{0.07} \\
Xtream (Ours) & \underline{0.16} & \underline{0.05} & \underline{0.05} & \textbf{0.08} & \underline{0.09} & \underline{0.08} \\
\midrule
\multicolumn{7}{l}{\cellcolor{gray!12}\textbf{Monocular-inertial methods}, ATE RMSE [m] $\downarrow$ after SE(3) alignment} \\
ORB-SLAM3 & 0.20 & 0.17 & 0.17 & 0.29 & 0.37 & 0.24 \\
MASt3R-Fusion (global) & \underline{0.13} & \underline{0.06} & 0.07 & \underline{0.12} & 0.09 & \underline{0.09} \\
MASt3R-Fusion & 0.19 & 0.27 & 0.21 & 0.36 & 0.19 & 0.24 \\
FoundationVIO$^\dagger$ & 0.19 & 0.07 & \underline{0.06} & 0.13 & \textbf{0.06} & 0.10 \\
Xtream (Ours) & \textbf{0.10} & \textbf{0.05} & \textbf{0.04} & \textbf{0.09} & \underline{0.07} & \textbf{0.07} \\
\bottomrule
\end{tabular*}}
\end{table}

\subsection{Ablation and hyperparameter studies}

Tables \ref{tab:ablation_vo} and \ref{tab:ablation_vi} show ablation studies of the main components of Xtream, for monocular and monocular-inertial setups, respectively. Explicitly modeling metric scale, our main contribution, yields by far the largest improvements, together with the inclusion of loop closures. 

Table~\ref{tab:ablation_handover} compares the scale handover process. Linking consecutive windows with Umeyama on the shared camera centres underperforms its alternatives. Depth links and prior metric information are complementary: the link constrains each pair of consecutive windows, giving local precision, while the prior anchors each window independently, bounding the drift that the chained links accumulate.

Table \ref{tab:abl_kf_count_outdoor} shows the impact of the keyframing policy on trajectory accuracy and computational cost. Reducing the number of keyframes per window to $8$ yields clear improvement in both metrics, with diminishing or less clear returns as the number of keyframes is reduced further. 

Table \ref{tab:loger_memory} reports trajectory accuracy metrics of Xtream using LoGeR* instead of DA3 as the 3DFM front end, in Oxford Spires. This shows how Xtream's gains hold for other 3DFMs. Further, we report results using LoGeR*'s memory, and with LoGeR*'s memory deactivated. In this last case, consistency between windows is achieved by geometric alignment. Results are similar in both cases, confirming that both strategies are prone to drift, and that, in contrast to LoGeR*'s implicit nature, Xtream's explicit metric representation brings again a substantial improvement, both in monocular and mono-inertial configurations.

\begin{table}[h!]
\centering
\footnotesize
\setlength{\aboverulesep}{0pt}\setlength{\belowrulesep}{0pt}\renewcommand{\arraystretch}{1.2}
\setlength{\tabcolsep}{3pt}
\caption{\textbf{Ablation study for monocular Xtream.} We report mean ATE [m] $\downarrow$ per dataset and relative improvement over plain 3DFM.}
\label{tab:ablation_vo}
\begin{tabularx}{\columnwidth}{Xrrrrr|r}
\toprule
Variant & KITTI & VBR & Oxford & EuRoC & ScaRF & Rel. \\
\midrule
3DFM streaming only & 29.7 & 72.0 & 14.4 & 0.29 & 0.85 & 1.00 \\
+ graph (depth link) & 29.5 & 72.1 & 14.5 & 0.29 & 0.87 & 1.00 \\
+ handover metric prior & 19.3 & 39.2 & 4.22 & 0.34 & 0.52 & 0.63 \\
+ covis.\ loops & 17.6 & 38.1 & 3.05 & 0.32 & 0.41 & 0.54 \\
+ loop closure (\textbf{ours}) & 9.79 & 5.40 & 1.67 & 0.07 & 0.08 & 0.16 \\
\bottomrule
\end{tabularx}
\end{table}

\begin{table}[h!]
\centering
\footnotesize
\setlength{\aboverulesep}{0pt}\setlength{\belowrulesep}{0pt}\renewcommand{\arraystretch}{1.2}
\setlength{\tabcolsep}{3pt}
\caption{\textbf{Ablation study for mono-inertial Xtream.} We report mean ATE [m] $\downarrow$ per dataset and relative improvement over 3DFM.}
\label{tab:ablation_vi}
\begin{tabularx}{\columnwidth}{Xrrrr|r}
\toprule
Variant & VBR & Oxford & EuRoC & ScaRF & Rel. \\
\midrule
3DFM streaming only$^\ast$ & 71.8 & 14.4 & 0.29 & 0.85 & 1.00 \\
+ graph (depth link, IMU) & 17.8 & 3.22 & 0.26 & 0.29 & 0.39 \\
+ covisibility loops & 17.1 & 2.36 & 0.24 & 0.27 & 0.33 \\
+ loop closures (\textbf{ours}) & 3.06 & 1.11 & 0.11 & 0.07 & 0.10 \\
\bottomrule
\end{tabularx}
\end{table}

\begin{table}[h!]
\centering
\footnotesize
\setlength{\aboverulesep}{0pt}\setlength{\belowrulesep}{0pt}\renewcommand{\arraystretch}{1.2}
\setlength{\tabcolsep}{3pt}
\caption{\textbf{Scale handover ablation (monocular).} ATE Sim(3) RMSE [m], mean per dataset; Rel.: geo-mean over all sequences of the ratio to the first row (streaming with loops, no handover).} 
\label{tab:ablation_handover}
\begin{tabularx}{\columnwidth}{Xrrrrr|r}
\toprule
Variant & KITTI & VBR & Oxford & EuRoC & ScaRF & Rel. \\
\midrule
None & 29.4 & 16.9 & 12.1 & 0.57 & 0.64 & 1.00 \\
Umeyama & 16.5 & 7.07 & 3.40 & 0.14 & 0.13 & 0.26 \\
Metric prior & 14.4 & 8.91 & 2.00 & 0.14 & 0.09 & 0.21 \\
Depth & 12.8 & 7.60 & 2.87 & 0.05 & 0.09 & 0.17 \\
Depth + m. prior (\textbf{ours}) & 9.79 & 5.40 & 1.67 & 0.07 & 0.08 & 0.15 \\
\bottomrule
\end{tabularx}
\end{table}

\begin{table}[t]
\centering
\footnotesize
\caption{\textbf{Impact of \# keyframes per window (mono-inertial).} ATE [m] $\downarrow$ after SE(3) and Sim(3) alignment and RPE [m/m] $\downarrow$.}
\label{tab:abl_kf_count_outdoor}
\setlength{\tabcolsep}{3pt}
\begin{adjustbox}{max width=\columnwidth}
\begin{tabular}{lrrrrrrrr}
\toprule
Keyframes & Frames & \multicolumn{3}{c}{\textbf{Oxford Spires}} & \multicolumn{3}{c}{\textbf{VBR}} & Back-end \\
per window & kept & SE3 & Sim3 & RPE & SE3 & Sim3 & RPE & time \\
\midrule
all (no keyframing) & 100\% & 2.262 & 1.170 & 0.049 & 5.532 & 3.114 & 0.107 & 1.00$\times$ \\
32 & 55\% & 2.096 & 1.091 & 0.048 & 4.836 & 2.976 & 0.101 & 0.62$\times$ \\
16 & 30\% & 1.452 & \textbf{0.795} & 0.045 & 3.775 & 2.616 & 0.092 & 0.39$\times$ \\
8 (\textbf{ours}) & 20\% & \textbf{1.107} & \underline{0.856} & \textbf{0.042} & 3.056 & \textbf{2.415} & \underline{0.087} & 0.26$\times$ \\
4 & 15\% & \underline{1.248} & 1.063 & \textbf{0.042} & \underline{2.757} & 2.478 & \underline{0.087} & \underline{0.21$\times$} \\
2 (first and last) & 10\% & 1.303 & 1.117 & \underline{0.043} & \textbf{2.695} & \underline{2.445} & \textbf{0.086} & \textbf{0.18$\times$} \\
\bottomrule
\end{tabular}
\end{adjustbox}
\end{table}

\begin{table}[h!]
\centering
\scriptsize
\setlength{\aboverulesep}{0pt}
\setlength{\belowrulesep}{0pt}
\renewcommand{\arraystretch}{1.2}
\setlength{\tabcolsep}{4pt}
\caption{\textbf{Xtream with LoGeR*'s memory activated (\cmark) and deactivated (\xmark), compared to LoGeR* alone.} We report ATE $\downarrow$, median scale drift $\downarrow$ after 1 and 10 windows, and RPE $\downarrow$.}
\label{tab:loger_memory}
\begin{tabular}{lccc}
\toprule
& \multicolumn{1}{c}{LoGeR*} 
& \multicolumn{1}{c}{\makecell{Xtream w/ LoGeR* \\ \textnormal{Monocular}}}
& \multicolumn{1}{c}{\makecell{Xtream w/ LoGeR* \\ \textnormal{Monocular-Inertial}}} \\
\cmidrule(lr){2-2}\cmidrule(lr){3-3}\cmidrule(lr){4-4}
Metric & Mem. \cmark / \xmark & Mem. \cmark / \xmark & Mem. \cmark / \xmark \\
\midrule
ATE Sim(3) [m]
& 6.2 / \textbf{6.1}
& 1.9 / \textbf{1.4}
& 0.9 / \textbf{0.7} \\

ATE SE(3) [m]
& --
& --
& 1.0 / \textbf{0.8} \\

Scale drift, 1 win. [\%]
& \textbf{2.1} / 3.3
& 1.5 / \textbf{1.3}
& \textbf{1.0} / \textbf{1.0} \\

Scale drift, 10 win. [\%]
& \textbf{11.8} / 12.0
& 4.4 / \textbf{3.6}
& 2.2 / \textbf{2.1} \\

RPE [cm/m]
& 17.1 / \textbf{17.0}
& 7.3 / \textbf{6.6}
& 3.9 / \textbf{3.8} \\
\bottomrule
\end{tabular}
\end{table}

\section{Conclusions}
\label{sec:conclusions}

In this paper we have presented Xtream, a hybrid pipeline for long-term streaming reconstruction from monocular and mono-inertial sequential inputs. Xtream combines a multi-view 3DFM front-end, which provides relative up-to-scale poses between sets of frames, with a factor graph that ensures local and global geometric consistency. Our main contribution is the explicit modeling of the metric state within the factor graph, through the introduction of scale ratios between the 3DFM scales and the metric scale. These ratios are updated in monocular setups by factors relating them to metric-scale measurements from metric-aware 3DFMs, and in monocular-inertial setups through IMU residuals. Our experiments demonstrate state-of-the-art performance across several outdoor benchmarks, where scale drift is the dominant source of trajectory error, in both monocular and monocular-inertial setups. Furthermore, Xtream achieves competitive results in small-scale scenarios, where scale drift is much less relevant.


{
    \small
    \bibliographystyle{ieeenat_fullname}
    \bibliography{main}
}

\clearpage
\setcounter{page}{1}
\maketitlesupplementary

\section*{A. Implementation details}
\label{sec:supp_impl}

Table \ref{tab:supp_params} summarizes the configuration for the main parameters of Xtream, further detailed below.

\noindent\textbf{Models.}
All multi-view passes (windows $\mathcal{W}_k$, covisibility loops $\mathcal{L}^{\mathrm{cov}}_{i,j}$ and closures
$\mathcal{L}^{\mathrm{lc}}_{i,j}$) run \emph{DA3-Giant} (DA3-GIANT-1.1) \cite{lin2025da3}. The monocular metric prior uses \emph{DA3-Metric-Large}, whose output is converted to metres with the focal length:
the calibrated one in the monocular-inertial setting, DA3-Giant's own per-keyframe prediction in the (uncalibrated) monocular setting.
Place recognition uses DINOv2 SALAD~\cite{izquierdo2024optimal}.

\noindent\textbf{Front end.}
Windows hold $W{=}64$ consecutive frames (no temporal subsampling) with an overlap of $O{=}3$ frames, and
the scale ratio $r_k$ of \cref{eq:ratio} is measured on the first shared frame. Each window contributes $K{=}8$
keyframes.

\noindent\textbf{Inertial noise.}
The IMU noise starts from each dataset's calibration, which is measured on a static sensor. In motion, vibration
and dynamics raise the noise well above these values, so we multiply the accelerometer noise density and both bias
random walks $(\sigma_{wa}, \sigma_{wg})$ by 10 and keep the gyroscope noise density, with the same factors on every
dataset. Each bias value $\bm{b}_\beta$ is shared by $R{=}5$ consecutive windows.

\noindent\textbf{Initialisation.}
In the monocular-inertial setting, $\ell_0$ is seeded by VI3's \cite{lozano2026vi3} velocity-scale initialisation over the IMU
samples of $\mathcal{W}_0$, as a prior with a standard deviation of 1 in log units.

\noindent\textbf{Solver and pruning.}
Levenberg--Marquardt starts with damping $10^{-5}$, takes one iteration per window update, and up to 100 iterations for loop updates and the final solve. Pruning blocks span 5 windows and keep the $K$ keyframes of each window.

\noindent\textbf{Dense map.}
We support both the base (default) streaming reconstruction mapping, and an extension of it with additional conditioned model passes. Map frames are sampled at 5\,Hz, back-projected every 4th pixel, and cut at a maximum range of 5\,m. The confidence threshold is the 25th percentile of the confidence of the pass that predicted the frame: the
window pass for the base map, the conditioned pass otherwise. Conditioned map passes take groups of 16 consecutive map frames with a stride of 12 (4 shared frames), with the input poses expressed in centimetres.

\noindent\textbf{Hardware.}
All experiments were run on a single NVIDIA RTX 5090 (32\,GB).

\begin{table}[t]
\centering
\footnotesize
\setlength{\tabcolsep}{3pt}
\caption{\textbf{Main parameters of Xtream.} The same values are used for all datasets.}
\label{tab:supp_params}
\begin{tabularx}{\columnwidth}{@{}c@{\hspace{4pt}}Xr@{}}
\toprule
& Parameter & Value \\
\midrule
\multirow{6}{*}{\rotatebox[origin=c]{90}{\scriptsize Front-end}}
& Window $W$ / overlap $O$ [frames] & 64 / 3 \\
& Keyframes per window $K$ & 8 \\
& SALAD threshold $\delta$ & 0.7 \\
& Covisibility search [windows] & 3 \\
& Closure separation [windows] & $\ge 4$ \\
& Closure candidates per query & top-5 \\
\midrule
\multirow{9}{*}{\rotatebox[origin=c]{90}{\scriptsize Back-end}}
& Visual factor $\sigma_R$ [rad] / $\sigma_t$ & 0.0025 / 0.02 \\
& Scale factor $\sigma_{\mathrm{s}}$ [log] & 0.08 \\
& Loop factor $\sigma^{\mathcal{L}}_R$ [rad] / $\sigma^{\mathcal{L}}_t$ & 0.01 / 0.1 \\
& Loop scale factor $\sigma^{\mathcal{L}}_{s}$ [log] & 0.05 \\
& Huber threshold (loops, metric prior) & 1.345 \\
& IMU noise factor (acc., gyro, walks) & 10 / 1 / 10 \\
& Student-$t$ $\nu$ / $\kappa$ / $p$ & 4.2 / 1.23 / 5 \\
& LM iterations (window / loop, final) & 1 / $\le 100$ \\
& Cond.\ map pass length / stride (opt.) & 16 / 12 \\
\bottomrule
\end{tabularx}
\end{table}

\section*{B. Additional results}
\label{sec:suppl_exps}

\noindent\textbf{Reconstruction.}
We provide results on the reconstruction task. Each map is scored considering that a dense map can fail in two ways:
its local geometry can be wrong, or drift can bend it out of shape over the whole run. Locally, the run is cut into
consecutive segments of 10\,m of camera travel; the sub-map of each segment is aligned on its own to the ground-truth
trajectory, refined with ICP against the ground-truth laser scan, and scored. The scores are averaged over
segments, removing accumulated drift and isolating the local geometry. Globally, the whole map is placed with a
single alignment to the ground-truth trajectory and no ICP, which also exposes trajectory drift. 

Each method is scored on its own map, built online: geometry is predicted once as the frames arrive and placed with the method's final
poses. The rest is common to all methods: map frames are sampled on the same 5\,Hz grid, points are cut at 5\,m from
the camera, and all maps are compared with the same reference cloud using the official evaluation tool.

We compare against two concurrent works: AMB3R-SLAM in the monocular setting and FoundationVIO in the mono-inertial
one, see Table \ref{tab:reconstruction_suppl}. Even without any extra network pass, Xtream's base map outperforms both on local and global F-score and on
accuracy. The conditioned map passes improve every one of these metrics further, and are presented as a higher quality add-on if increased cost is allowed (see Table \ref{tab:runtime}). Completeness is the exception, where AMB3R-SLAM comes out ahead. Unlike recall, completeness rewards points that are spread out regardless of their
accuracy, and on recall Xtream is clearly ahead.

\begin{table}[ht]
\centering
\footnotesize
\setlength{\aboverulesep}{0pt}\setlength{\belowrulesep}{0pt}\renewcommand{\arraystretch}{1.2}
\setlength{\tabcolsep}{3pt}
\caption{Reconstruction on Oxford Spires, mean over the 11 sequences. Local: precision, recall and F-score [\%] at 5\,cm after per-chunk alignment (10\,m + ICP). Global: F-score at 5 and 10\,cm, accuracy and completeness [m] after one Sim(3) (monocular) or SE(3) (mono-inertial) alignment. Base: depth from the front end's window passes; + cond.\ map passes: extra 3DFM passes conditioned on the live poses (cost in Table~\ref{tab:runtime}). Best in \textbf{bold} per group.}
\label{tab:reconstruction_suppl}
\begin{tabularx}{\columnwidth}{Xrrrrrrr}
\toprule
& \multicolumn{3}{c}{Local} & \multicolumn{4}{c}{Global} \\
\cmidrule(lr){2-4}\cmidrule(lr){5-8}
Method & P\ensuremath{\uparrow} & R\ensuremath{\uparrow} & F@5\ensuremath{\uparrow} & F@5\ensuremath{\uparrow} & F@10\ensuremath{\uparrow} & Acc.\ensuremath{\downarrow} & Comp.\ensuremath{\downarrow} \\
\midrule
\multicolumn{8}{l}{\textit{Monocular}} \\
AMB3R-SLAM & 34.0 & 42.7 & 36.4 & 11.9 & 25.1 & 0.53 & \textbf{3.69} \\
Xtream (base) & 44.8 & 56.7 & 48.7 & 18.4 & 34.3 & 0.41 & 4.10 \\
\quad + cond.\ pass & \textbf{49.1} & \textbf{59.3} & \textbf{52.4} & \textbf{20.3} & \textbf{36.8} & \textbf{0.39} & 4.08 \\
\midrule
\multicolumn{8}{l}{\textit{Mono-inertial}} \\
FoundationVIO & 41.2 & 58.0 & 45.7 & 18.2 & 34.4 & 0.39 & 4.49 \\
Xtream (base) & 46.5 & 57.4 & 50.0 & 21.7 & 39.9 & 0.29 & \textbf{4.37} \\
\quad + cond.\ pass & \textbf{55.9} & \textbf{63.8} & \textbf{58.0} & \textbf{24.5} & \textbf{45.1} & \textbf{0.28} & 4.43 \\
\bottomrule
\end{tabularx}
\end{table}

\noindent\textbf{Runtime.}
Table~\ref{tab:runtime} reports Xtream's runtime on a single GPU, timed during the full runs with every component
included. The 3DFM window pass dominates the cost. All other work runs alongside the stream: covisibility edges, loop
closures and the back-end solve. When the last frame arrives, only a fraction of a second of back-end work remains,
so the poses are final almost immediately. Covisibility edges are built only where windows actually see the same
scene, so their cost follows the motion. On KITTI, the car leaves each window's view within seconds and few edges form,
which makes the system fastest there. The conditioned map passes are optional: they lower throughput moderately even
when counted as run after the stream, and memory stays below the run's peak.

\begin{table}[t]
\centering
\footnotesize
\setlength{\aboverulesep}{0pt}\setlength{\belowrulesep}{0pt}\renewcommand{\arraystretch}{1.2}
\setlength{\tabcolsep}{3pt}
\caption{Runtime of Xtream on one RTX 5090, median over the sequences of each dataset. FPS: input frames over the wall time of the whole run, everything included. Window pass: one 3DFM call ($W-O = 61$ new frames). Back-end drain: time left after the last frame. Shares: per-component time over wall time, monocular runs (mono-inertial within 2 points, without the metric prior). KITTI has no IMU (--). $^\dagger$Monocular; mono-inertial needs 1--1.3\,GB less. $^\ddagger$With the conditioned map passes, counted as run after the stream (an upper bound on their cost).}
\label{tab:runtime}
\begin{tabularx}{\columnwidth}{Xrrrrr}
\toprule
& KITTI & VBR & Oxford & EuRoC & ScaRF \\
\midrule
FPS, monocular & 21.0 & 14.3 & 10.8 & 12.0 & 9.9 \\
\quad + cond.\ map passes$^\ddagger$ & 14.0 & 11.7 & 9.2 & 10.3 & 8.9 \\
FPS, mono-inertial & -- & 12.2 & 11.5 & 12.1 & 9.3 \\
\quad + cond.\ map passes$^\ddagger$ & -- & 10.5 & 9.4 & 10.4 & 8.5 \\
\midrule
Peak GPU memory [GB]$^\dagger$ & 13.1 & 12.5 & 15.4 & 14.1 & 15.4 \\
\midrule
Window pass [s] & 2.28 & 1.98 & 3.44 & 2.75 & 3.42 \\
Back-end drain [s] & 0.06 & 0.17 & 0.11 & 0.06 & 0.07 \\
\midrule
\multicolumn{6}{l}{\textit{Share of wall time [\%]}} \\
3DFM window pass & 82 & 46 & 63 & 54 & 56 \\
Covisibility edges & 2 & 14 & 16 & 12 & 13 \\
Place recognition & 8 & 15 & 6 & 5 & 4 \\
Loop closure & 5 & 21 & 7 & 31 & 28 \\
Metric prior & 3 & 3 & 3 & 2 & 2 \\
\bottomrule
\end{tabularx}
\end{table}

\noindent\textbf{Metric scale.}
Table~\ref{tab:scale_drift} separates two properties of scale: accuracy, i.e. whether the trajectory has the right
metric size overall, and drift, i.e. whether that scale stays stable along the run. With the IMU, Xtream recovers physically accurate metric scale on every dataset. The learned metric prior used in monocular vision also yields metric trajectories, but with a clearly larger global error. Scale drift is low in both settings, since every window is tied to its predecessor
through the shared frames, and the IMU reduces it further on most datasets. EuRoC is an exception, where the two settings are comparable. Its short sequences in small rooms leave little room for scale to drift, and therefore for the IMU to correct. The IMU's benefit grows on the long, large-scale runs of VBR and Oxford Spires. These are precisely the settings that challenge long-term consistency, and where Xtream excels.

\begin{table}[!tbp]
\centering
\footnotesize
\setlength{\aboverulesep}{0pt}\setlength{\belowrulesep}{0pt}\renewcommand{\arraystretch}{1.2}
\setlength{\tabcolsep}{3pt}
\caption{Scale of Xtream's trajectories, mean over the sequences of each dataset. Metric scale error: $|s-1|$ of a Sim(3) fit of the whole trajectory to the ground truth. Scale drift: median scale change between front-end windows 10 apart ($\approx$1 min), each scale from a Sim(3) fit of the window's own segment (segments under 0.5 m skipped). Monocular scale comes from the metric depth prior, mono-inertial from the IMU. Best per column in \textbf{bold}.}
\label{tab:scale_drift}
\begin{tabularx}{\columnwidth}{Xrrrrr}
\toprule
 & KITTI & VBR & Oxford & EuRoC & ScaRF \\
\midrule
\multicolumn{6}{l}{\textit{Metric scale error [\%] \ensuremath{\downarrow}}} \\
Monocular & 7.3 & 10.9 & 13.7 & 23.1 & 7.1 \\
Mono-inertial & -- & \textbf{1.7} & \textbf{1.5} & \textbf{1.0} & \textbf{0.6} \\
\midrule
\multicolumn{6}{l}{\textit{Scale drift over 10 windows [\%] \ensuremath{\downarrow}}} \\
Monocular & 4.0 & 6.6 & 5.5 & \textbf{2.5} & 2.7 \\
Mono-inertial & -- & \textbf{3.6} & \textbf{2.8} & 3.2 & \textbf{1.6} \\
\bottomrule
\end{tabularx}
\end{table}

\noindent\textbf{Loop gating.}
Gating unreliable loops, Tab.~\ref{tab:ablation_gate}, matters on long runs, where wrong loops are frequent. Rejection only targets gross outliers: moderately wrong loops
pass the test and enter the graph at full weight. Down-weighting handles these too, and is the better choice when the
IMU provides a reliable reference for each loop's innovation.

\begin{table}[t]
\centering
\footnotesize
\setlength{\aboverulesep}{0pt}\setlength{\belowrulesep}{0pt}\renewcommand{\arraystretch}{1.2}
\setlength{\tabcolsep}{3pt}
\caption{Loop gating ablation: ATE RMSE [m], mean per dataset. Rel.: geometric mean over all sequences of the ratio to no gate in that mode. Weight: Student-t weight before insertion, from the loop's innovation against the graph, its scale from the window vision factors (mono-inertial) or from earlier loops' innovations (monocular). Reject: loop rejection.}
\label{tab:ablation_gate}
\begin{tabularx}{\columnwidth}{Xrrrrr|r}
\toprule
\textit{Monocular} & KITTI & VBR & Oxford & EuRoC & ScaRF & Rel. \\
\midrule
no gating & 9.79 & 8.77 & 1.64 & 0.07 & 0.08 & 1.00 \\
weight & 10.1 & 7.70 & 1.44 & 0.09 & 0.09 & 1.03 \\
reject (\textbf{ours}) & 9.79 & 5.40 & 1.67 & 0.07 & 0.08 & 0.95 \\
\bottomrule
\end{tabularx}
\vspace{4pt}
\begin{tabularx}{\columnwidth}{Xrrrr|r}
\toprule
\textit{Mono-inertial} & VBR & Oxford & EuRoC & ScaRF & Rel. \\
\midrule
no gating & 4.46 & 1.28 & 0.11 & 0.07 & 1.00 \\
reject & 3.11 & 1.31 & 0.11 & 0.07 & 0.97 \\
weight (\textbf{ours}) & 3.06 & 1.11 & 0.11 & 0.07 & 0.94 \\
\bottomrule
\end{tabularx}
\end{table}

\end{document}